\documentclass[10pt,twocolumn,letterpaper]{article}
\usepackage{cvpr}
\usepackage{times}
\usepackage{epsfig}
\usepackage{graphicx}
\usepackage{amsmath}
\usepackage{amssymb}
\usepackage{dsfont}
\usepackage{nccmath}
\usepackage{bm}
\usepackage{bbm}
\usepackage{threeparttable}
\usepackage{booktabs}
\usepackage{algorithm}
\usepackage{algorithmic}
\usepackage{tabularx}
\usepackage{pifont}

\newcommand{\cmark}{\ding{51}}%
\newcommand{\xmark}{\ding{55}}%

\usepackage[breaklinks=true,bookmarks=false]{hyperref}

\cvprfinalcopy 

\def\cvprPaperID{4330} 

\ifcvprfinal\fi
\begin{document}

\title{Learning Event-Based Motion Deblurring}
\author{Zhe Jiang\textsuperscript{\thanks{Equal contribution. This work is done when Jiang Zhe is interned with SenseTime Research.} \ 1,2} \quad Yu Zhang\textsuperscript{*\thanks{Correspondence should be addressed to: \url{zhangyulb@gmail.com}} \ 1,3} \quad Dongqing Zou\textsuperscript{1} \quad Jimmy Ren\textsuperscript{1} \quad Jiancheng Lv\textsuperscript{2} \quad Yebin Liu\textsuperscript{3}\\
\textsuperscript{1}SenseTime Research \quad \textsuperscript{2}Sichuan University \quad \textsuperscript{3}Department of Automation, Tsinghua University
}

\maketitle

\begin{abstract}
   Recovering sharp video sequence from a motion-blurred image is highly ill-posed due to the significant loss of motion information in the blurring process. For event-based cameras, however, fast motion can be captured as events at high time rate, raising new opportunities to exploring effective solutions. In this paper, we start from a sequential formulation of event-based motion deblurring, then show how its optimization can be unfolded with a novel end-to-end deep architecture. The proposed architecture is a convolutional recurrent neural network that integrates visual and temporal knowledge of both global and local scales in principled manner. To further improve the reconstruction, we propose a differentiable directional event filtering module to effectively extract rich boundary prior from the stream of events. We conduct extensive experiments on the synthetic GoPro dataset and a large newly introduced dataset captured by a DAVIS240C camera. The proposed approach achieves state-of-the-art reconstruction quality, and generalizes better to handling real-world motion blur.
\end{abstract}

\vspace{-1mm}
\section{Introduction}
Motion blur happens commonly due to the exposure time required by modern camera sensors, during which scenes are recorded at different time stamps and accumulated into averaged (blurred) signal. The inverse problem called \textit{deblurring}, which unravels the underlying scene dynamics behind a motion-blurred image and generates a sequence of sharp recovery of the scene, is still challenging in computer vision. While simple motion patterns (\eg~camera shake) have been well modelled~\cite{PanCVPR16,MichaeliECCV14,ChakrabartiECCV16,DongICCV17,YanCVPR17,HirschICCV11,ZhangNIPS13,BahatICCV17}, formulating more sophisticated motion patterns in real world, however, is much more difficult.

\begin{figure}[!t]
	\centering
	\includegraphics[width=1.0\columnwidth]{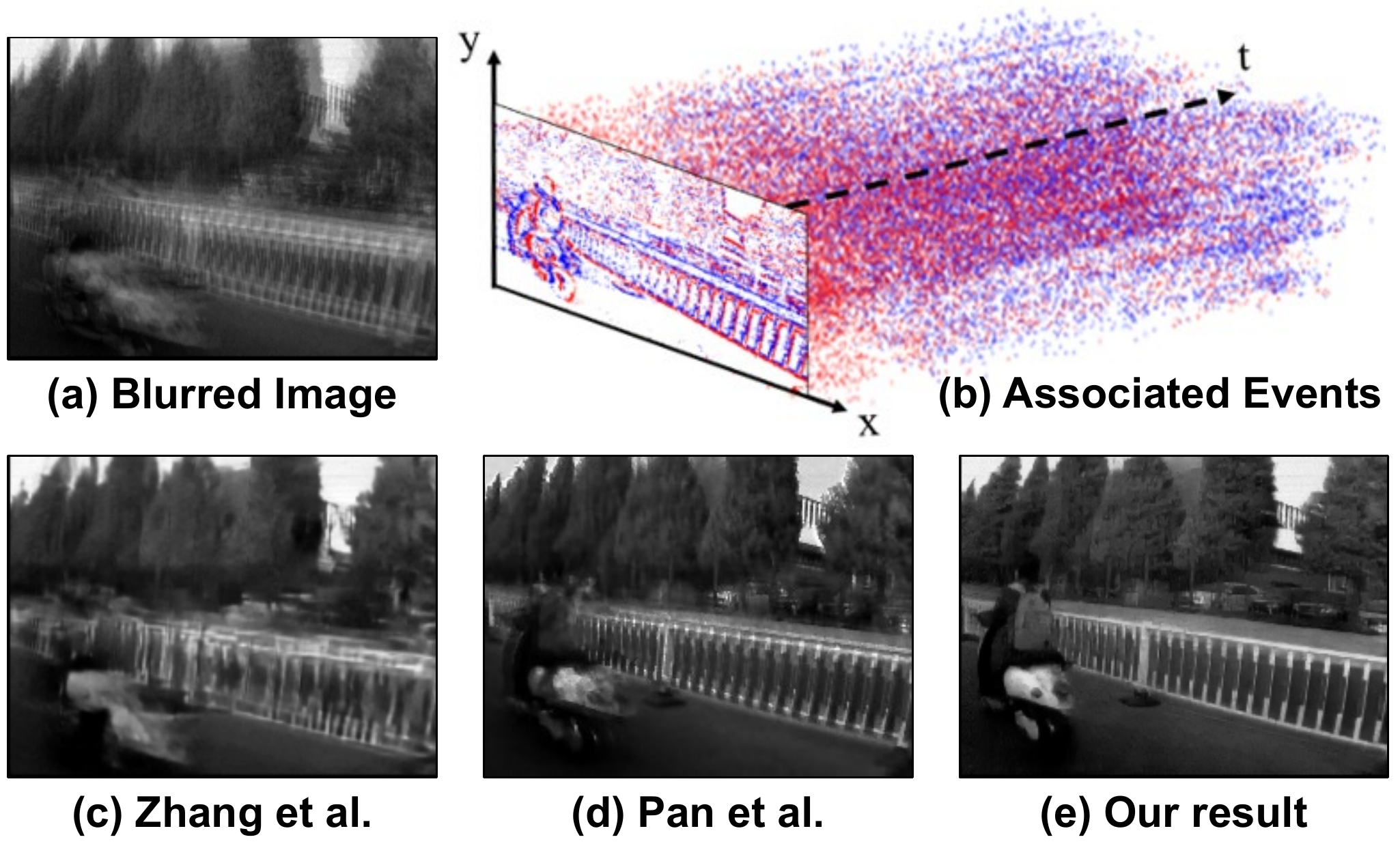}
	\caption{Motivation of our approach. A severe motion-blurred image (a) is difficult to deblur by observing its ambiguous appearance only even with state-of-the-art deep architecture~\cite{ZhangCVPR19} (c). Though events (b) provide dense temporal cues, the physical reconstruction approach~\cite{PanCVPR19} still presents unaddressed blur due to the noisiness of events (d). The proposed deep motion deblurring learns to recover plausible details from imperfect image and events (e).}
	\label{fig:intro}
	\vspace{-3mm}
\end{figure}

To model general motion blur, recent deep learning approaches propose to recover a blurred image by observing lots of sharp images and their blurred versions~\cite{SunCVPR15,GongCVPR17,ZhangCVPR18,NahCVPR17,JinCVPR18,TaoCVPR18}. Despite their success in certain scenarios, they may fail reconstructing the scene plausibly for severe motion blur (\eg~Fig.~\ref{fig:intro}), which is common for handheld, vehicle or drone-equipped cameras. In this case, hallucinating the scene details is hardly possible due to the significant loss of temporal order and visual information.

Instead of purely relying on computational architectures, this work adopts event-based cameras to alleviate this problem at data capture stage. Event cameras are biologically inspired sensors adept at recording the change of pixel intensities (called \textit{events}) with microsecond accuracy and very low power consumption. The hybrid model of such sensors (\eg~\cite{BrandliJSSC14}) allows the events being temporally calibrated with the image. As a result, such data naturally encodes dense temporal information that can facilitate motion deblurring. As shown in Fig.~\ref{fig:intro} (a) and (b), although the image undergoes significant blur, the accompanying events are temporally dense and reveal clear moving pattern of the scene.

Despite the high potential of event-based motion deblurring, a critical issue is that events are lossy and noisy signals triggered only if pixel intensity changes up to certain threshold that can vary with the change of scene conditions~\cite{RebecqCORL18}. Such discrete and inconsistent sampling makes textures and contrast difficult to restore. As shown in Fig.~\ref{fig:intro} (d), state-of-the-art physical deblurring method~\cite{PanCVPR19} still has difficulty reconstructing the image plausibly. Our solution is to plug deeply learned priors into event-based deblurring process, so as to surpass the imperfectness of data.

In details, this work starts from a sequential formulation of event-based deblurring. By reinterpreting its optimization with deep networks, we propose a novel recurrent architecture trainable end-to-end. For each time step, coarse reconstructions are obtained from previous reconstruction as well as the local temporal events. Fine details are then supplied by network predictions, guided by appearance and temporal cues at both global and local scales. To further improve the quality of reconstruction, we propose a differentiable Directional Event Filtering (DEF) module, which effectively aggregates the motion boundaries revealed by events and produces sharp deblurring prior. To evaluate the proposed approach, we compile a large outdoor dataset captured using the DAVIS240C camera~\cite{BrandliJSSC14}. Extensive experiments on this dataset and the synthetic GoPro dataset~\cite{NahCVPR17} show that the proposed approach outperforms various state-of-the-art methods, either image-based or event-based, and generalizes better to handling real-world motion blur.

Contributions of this paper are summarized as follows. 1) We propose a novel recurrent deep architecture for event-based motion deblurring, which achieves state-of-the-art results on two large benchmarks. 2) We propose directional event filtering to generate sharp boundary prior from events for motion deblurring. 3) We compile a new event dataset with real-world motion blur to facilitate future research.

\section{Related Work}
\textbf{Blind motion deblurring} aims to resolve a blurry image without knowing the blurring kernel. Early works have designed various blurring-aware indicators, such as color channel statistics~\cite{PanCVPR16,YanCVPR17}, patch recurrence~\cite{MichaeliECCV14} and ``outlier'' image signals~\cite{DongICCV17}, to define latent image priors. Several works propose to learn motion kernels~\cite{SchmidtCVPR13,PanCVPR17}, restoration functions~\cite{XiaoECCV16,GongCVPR17} and image priors~\cite{ZuoCVPR15,SunCVPR15} from data. More complex motion patterns compounded by different objects were also addressed~\cite{KimICCV13,RenICCV17}. Richer prior knowledge such as scene geometry was proven useful~\cite{PanLCVPR17,ParkICCV17}.

A recent trend is to approach all the complexities of motion deblurring with deep neural networks. Various kinds of effective network designs are proposed, including enlarging the receptive field~\cite{ZhangCVPR18}, multi-scale fusion~\cite{NahCVPR17,NorooziGCPR17}, feature distangling~\cite{NimishaICCV17}, and recurrent refinement~\cite{WieschollekICCV17}. There was also research on decoding the motion dynamics of a blurred image to a sharp video sequence~\cite{JinCVPR18}. Despite these advances, the considerable combinations of real-world lightings, textures and motions, which are severely missing in a blurred image, are still difficult to be plausibly recovered.

\textbf{Event cameras}~\cite{LichtsteinerJSSC08,BrandliJSSC14} are a special kind of sensors that detect intensity changes of the scene at microsecond level with slight power consumption. They find applications in various vision tasks, such as visual tracking~\cite{RameshBMVC18,MitrokhinFPA18}, stereo vision~\cite{ZhuECCV18, AndreopoulosCVPR18} and optical flow estimation~\cite{LiuBMVC18,YeArXiv18}. A related branch is to explore the corrupted event signals to restore high frame rate image sequences~\cite{ScheerlinckACCV18,MundaIJCV18,ShedligeriArXiv18}. Recently, Pan~\etal~\cite{PanCVPR19} formulates event-based motion blurring with a double integral model. Yet, the noisy hard sampling mechanism of event cameras often introduces strong accmulated noise and loss of scene details/contrast. 

This work shares the insight of recent works on event-to-video translation~\cite{PiniArXiv18,MostafaviCVPR19,RebecqCVPR19} that surpasses the imperfect event sampling by learning plausible details from data. While~\cite{PiniArXiv18} addresses future frame prediction,~\cite{MostafaviCVPR19,RebecqCVPR19} translate events to plausible intensity images in streaming manner depending on local motion cues. Instead, this work explores both long-term, local appearance/motion cues as well as novel event boundary priors to solve motion deblurring.

\section{Learning Event-Based Motion Deblurring}
\label{sec:formulation}
Given a motion-blurred image $\bar{\mathcal{I}}$, our objective is to recover a sharp video sequence with $T$ frames, $\mathbb{I}=\{ \mathcal{I}_i \}_{i=1}^T$. We assume that a set of events $\mathbb{E}_{1 \sim T}$ are also captured by hybrid image-event sensors during the exposure, where the tilde denotes the time interval. Each event $\mathcal{E} \in \mathbb{E}_{1 \sim T}$ has the form $\mathcal{E}_{x, y, t}$, meaning that it is triggered at image coordinate $\left( x, y \right)$ and time point $t \in \left[1, T \right]$. Note here $t$ does not need to be an integer, but can be fractional due to the high temporal resolution (\ie microsecond-level) of event camera. A polarity $p_{x,y,t}$ is recorded for $\mathcal{E}_{x,y,t}$ indicating the change of local intensity. Formally, it is defined as~\cite{LichtsteinerJSSC08,BrandliJSSC14}
\begin{equation}
p_{x,y,t} = 
\begin{cases}
+1, \text{if} \log \left( \frac{\mathcal{I}_t {(x, y)}} {\mathcal{I}_{ t - \Delta t} {(x, y)} } \right) > \tau, \\
-1, \text{if} \log \left( \frac{\mathcal{I}_t {(x, y)}} {\mathcal{I}_{ t - \Delta t} {(x, y)} } \right) < -\tau, 
\end{cases}
\label{eq:event}
\end{equation}

Eqn.~\eqref{eq:event} shows that, events are triggered if the instant image at time point $t$, namely $\mathcal{I}_t$, has pixel intensity changed up to a threshold $\pm\tau$ in a small time period $\Delta t$. Without loss of generality, we assume that $p_{x,y,t}$ takes zero in case that $\log \left( \frac{\mathcal{I}_t {(x, y)}} {\mathcal{I}_{ t - \Delta t} {(x, y)} } \right)$ is in $\left[-\tau, \tau \right]$. For adjacent latent images $\mathcal{I}_i$ and $\mathcal{I}_{i-1}$, the following relationship can be derived:
\begin{equation}
\mathcal{I}_{i} \left(x, y \right) \approx \mathcal{I}_{i-1} \left(x, y \right) \cdot \exp \left( \tau \int_{t=i-1}^{i} p_{x,y,t} \mathbbm{1} \left(\mathcal{E}_{x,y,t} \right) dt \right),
\label{eq:event_update}
\end{equation}
The indicator function $\mathbbm{1} \left( \cdot \right)$ equals $1$ if the event $\mathcal{E}_{x,y,t}$ exists, or $0$ otherwise.

One should note that the approximation error of~\eqref{eq:event_update} is getting lower when $\Delta t, \tau \to 0$, which implies denser events according to~\eqref{eq:event}. However, with inconsistent $\tau$ affected by various kinds of noise, the approximation is mostly insufficient in practice, leading to loss of contrast and details. To address this issue, we propose a joint framework that learns to reconstruct clean images from data, by reinterpreting a sequential deblurring process.

\textbf{Deep sequential deblurring.} Event-assisted deblurring can be formulated under Maximum-a-Posteriori:
\begin{equation}
\mathbb{I}^* = \arg\max_{\mathbb{I}} P \left( \mathbb{I} | \bar{\mathcal{I}}, \mathbb{E}_{1 \sim T} \right).
\label{eq:map}
\end{equation}
To solve the combinatorial problem~\eqref{eq:map} we make the following simplifications. For the joint posterior $  P \left( \mathbb{I} | \bar{\mathcal{I}}, \mathbb{E}_{1 \sim \mathcal{T}} \right) $, we make use of the temporal relations between adjacent latent images~\eqref{eq:event_update}, and assume a Markov chain model:
\begin{equation}
\begin{split}
 P \left( \mathbb{I} | \bar{\mathcal{I}}, \mathbb{E}_{1 \sim T} \right) \approx 
 & P \left( \mathcal{I}_T | \bar{\mathcal{I}}, \mathbb{E}_{1 \sim T} \right) \times \\ 
 & \prod_{i=1}^{T-1} P \left( \mathcal{I}_i | \mathcal{I}_{i+1}, \bar{\mathcal{I}}, \mathbb{E}_{1 \sim T} \right),
\end{split}
\label{eq:markov}
\end{equation}
in which $P \left( \mathcal{I}_i | \mathcal{I}_{i+1}, \bar{\mathcal{I}}, \mathbb{E}_{1 \sim T} \right) = P \left( \mathcal{I}_i | \mathcal{I}_{i+1}, \bar{\mathcal{I}}, \mathbb{E}_{i \sim i+1} \right)$ with Markov assumption. Note that this simplified model first estimates $\mathcal{I}_T$, then perform sequential reconstruction in backward order. According to Bayesian rule, the maximizer of a backward reconstruction step equals to:
\begin{equation}
\mathcal{I}^{*}_i = \arg\max_{\mathcal{I}_i} P \left( \mathcal{I}_{i+1}, \bar{\mathcal{I}}, \mathbb{E}_{i \sim i+1} | \mathcal{I}_i \right) P \left( \mathcal{I}_i \right).
\label{eq:seq_model}
\end{equation} 
Here, the prior term $P \left( \mathcal{I}_i \right)$ imposes desired distributions of the latent image, \eg~$\ell_1$ gradient~\cite{BardowCVPR16} or manifold smoothness~\cite{MundaIJCV18} in recent event-based image reconstruction. To model the likelihood term, we assume that there is an initial estimate from previous reconstruction, via~\eqref{eq:event_update}:
\begin{equation}
\hat{\mathcal{I}}_{i} = \mathcal{I}_{i+1} \odot \exp \left( - \tau \mathcal{S}_{i+1}^{i} \right),
\label{eq:event_update2}
\end{equation}
where $\forall x, y, \mathcal{S}_{i+1}^{i} \left( x, y \right) =\int_{t=i}^{i+1} p_{x,y,t} \mathbbm{1} \left(\mathcal{E}_{x,y,t} \right) dt $, and $\odot$ denotes Hadamard product. Since the time interval is small, we assume constant $\tau$ which introduces only small drift and provides good initialization. To solve $\mathcal{I}^*_i$, several works assume simple distributions centered around $\hat{\mathcal{I}}^*_i$ to define the likelihood term in~\eqref{eq:seq_model}, \eg in~\cite{MundaIJCV18} a Poisson distribution is used. In this manner, Eqn.~\eqref{eq:seq_model} can be treated as a well-studied denoising problem.

Instead of using simple image prior, we borrow from recent research on learning deep denoising prior~\cite{ZhangICCV17,ZhangCVPR19}. In particular, we plug a deep network $\mathcal{N}$ as a learned denoiser,
\begin{equation}
\mathcal{I}^*_i = \mathcal{N} \left( \hat{\mathcal{I}}_i, \mathcal{I}_{i+1}, ,  \bar{\mathcal{I}}, \mathbb{E}_{i \sim i+1} \right).
\label{eq:denoising}
\end{equation}
As such, prior of latent image $P \left( \mathcal{I}_i \right)$ is not explicitly defined but implicitly learned from training data. To reduce parameter size and prevent overfitting, we use the same network governed by the same set of parameters for each deblurring step of~\eqref{eq:seq_model}, leading to a recurrent architecture.

\begin{figure*}[!t]
	\centering
	\includegraphics[width=1.0\textwidth]{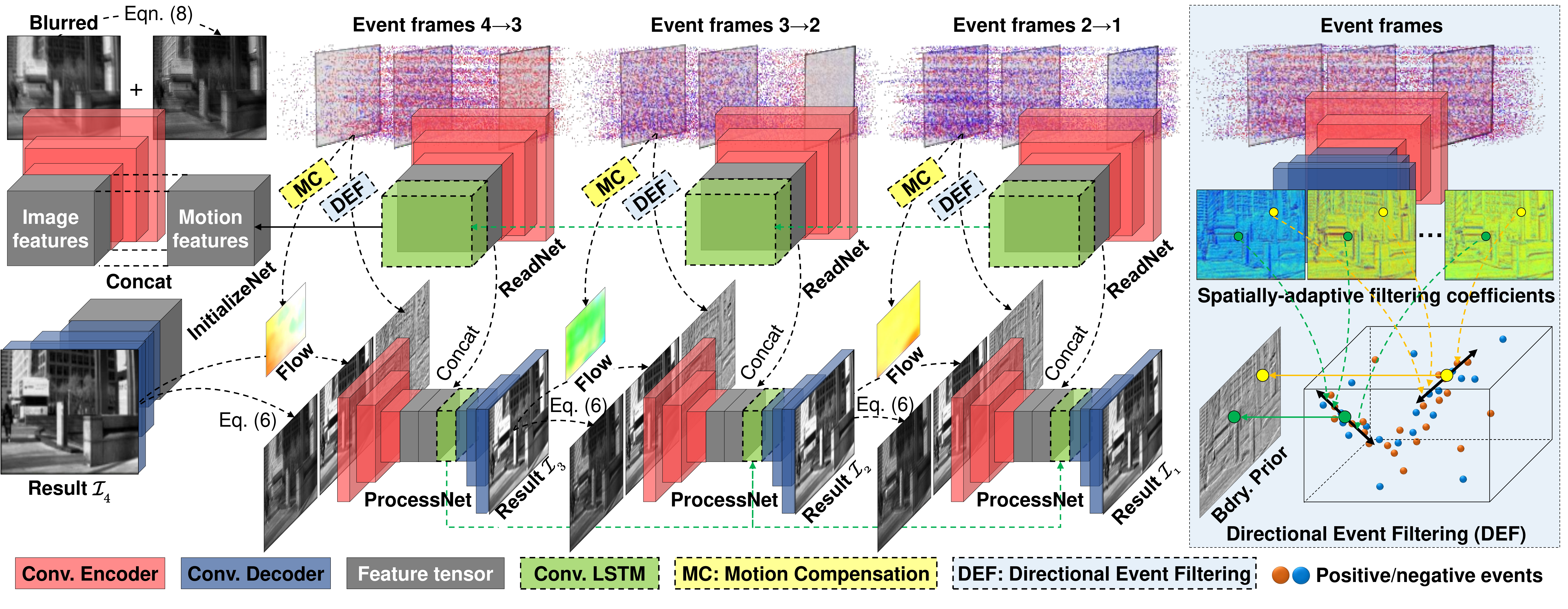}
	\caption{The proposed learning framework for event-based motion deblurring. For better visualization, we only assume $4$ sharp frames are recovered from the blurred image. Detailed layer and parameter configurations are referred to the supplementary material. Note that the Motion Compensation (MC) module is not illustrated due to the lack of space. See text for detailed description of the architecture.  } 
	\label{fig:pipeline}
\end{figure*}

The remaining problem of solving~\eqref{eq:markov} is how to get the initial latent image, \ie~$\mathcal{I}_T$. We use the fact that the blurred image $\bar{\mathcal{I}}$ roughly equals the average of the instant images in the exposure process. Combining this fact with~\eqref{eq:event_update2}, we have
\begin{equation}
\bar{\mathcal{I}} \approx \frac{1}{T} \sum_{i=1}^T{\mathcal{I}_i} =  \mathcal{I}_{T} \odot \frac{1}{T} \left( 1 + \sum_{t=2}^T \prod_{i=1}^{t-1} \mathcal{B}_{T - i + 1} ^ {T - i} \right),
\label{eq:solve_init}
\end{equation}
where $\mathcal{B}_{i+1}^i = \exp \left( - \tau \mathcal{S}_{i+1}^i \right) $ and $\mathcal{S}_{i+1}^i$ is defined in~\eqref{eq:event_update2}. It provides an initial estimation of $\mathcal{I}_T$, namely $\hat{\mathcal{I}}_T$, using the blurred image $\bar{\mathcal{I}}$ and events. Thus, we also treat solving $\mathcal{I}_T$ a denoising problem, centered around $\hat{\mathcal{I}}_T$, and use a network to approximate it. We note, however, the accumulative operator in~\eqref{eq:solve_init} introduces more drift unlike the sequential deblurring steps. We thus correct $\hat{\mathcal{I}}_T$ via a separate and more powerful network: $
\mathcal{I}^*_T = \mathcal{N}_0 \left( \hat{\mathcal{I}}_T, \bar{\mathcal{I}}, \mathbb{E}_{1 \sim T} \right) $. The full deblurring process is summarized in Alg.~\ref{alg:alg}. Note that by design~\eqref{eq:denoising}, the latent image is conditioned on both local and long-term cues from the image and events.
\begin{algorithm}
	\caption{Event-assisted Deep Motion Deblurring}
	\label{alg:alg}
	\begin{algorithmic}[1]
		\REQUIRE the blurred image $\bar{\mathcal{I}}$, events $\mathbb{E}_{1 \sim T}$
		\STATE Get initial estimate $\hat{\mathcal{I}}_T$ by solving~\eqref{eq:solve_init}
		\STATE Deblurring: $\mathcal{I}^*_T = \mathcal{N}_0 \left( \hat{\mathcal{I}}_T, \bar{\mathcal{I}}, \mathbb{E}_{1 \sim T} \right) $
		\STATE Initialize counter: $ i = T - 1 $
		\WHILE{$i \geq 1$}
		\STATE Get initial estimate $\hat{\mathcal{I}}_i$ by solving~\eqref{eq:event_update}
		\STATE Deblurring: $ \mathcal{I}^*_i = \mathcal{N} \left( \hat{\mathcal{I}}_i, \mathcal{I}_{i+1}, \mathbb{E}_{i \sim i+1},  \bar{\mathcal{I}}, \mathbb{E}_{1 \sim T} \right) $
		\STATE $i \leftarrow i - 1$
		\ENDWHILE
		\RETURN Deblurred sequence $ \mathbb{I}^* = \{ \mathcal{I}_i \}_{i=1}^T$
	\end{algorithmic}
\end{algorithm}

\vspace{-2mm}
\section{Network Architecture}
Fig.~\ref{fig:pipeline} shows the proposed event-based motion deblurring architecture, which contains: a \textit{read} network that traverses over the events and generates a single representation of the global scene motion, an \textit{initialize} network that couples appearance and motion to generate the initial latent image, and the recurrent \textit{process} network sequentially deblurring all the latent images\footnote{Due to space limit we briefly describe the component design and refer the detailed layer/parameter configurations to our supplementary material.}. The read and initialize networks instantiates $\mathcal{N}_0$ while the process network implements $\mathcal{N}$ in Alg.~\ref{alg:alg}.

The read network reads all the event data and generate a joint representation that accounts for the global event motion. To accomplish that, events during the exposure are first binned into equal-length time intervals ($3$ intervals in Fig.~\ref{fig:pipeline}). In each time interval, events are represented with stacked event frames~\cite{MostafaviCVPR19}, through further dividing an interval into $8$ equal-size chunks, summing over the polarities of events falling into each chunk, and stacking the results along channel dimension. The read network is a recurrent encoder consisting of convolutional blocks and a convolutional LSTM~\cite{ShiNIPS15} on top to accumulate features along time.

The initialize network decodes the appearance from the blurred image and couples it with the global motion to solve the latent image $\mathcal{I}^*_T$. It takes as input both the blurred image $\bar{\mathcal{I}}$ and the initial estimate $\hat{\mathcal{I}}_T$ (via solving Eqn.~\eqref{eq:solve_init}) and processes them with a convolutional encoder, concatenates the encodings with the accumulated global motion features from the read network, and feeds the joint features into a decoder to get the result.

Given the initial result, the process network then sequentially deblurs the remaining latent images. In the $i$th step, it consumes both image and event-based observations. The image part include: 1) the initial estimate $\hat{\mathcal{I}}_i$ as obtained by Eqn.~\eqref{eq:event_update2} using the previous reconstruction $\mathcal{I}_{i+1}$, 2) the local historical image by transforming the previous result $\mathcal{I}_{i+1}$ with the \textit{Motion Compensation} (``MC'' in Fig.~\ref{fig:pipeline}) module, and 3) the boundary guidance map given by the \textit{Directional Event Viltering} (``DEF'' in Fig.~\ref{fig:pipeline}) module. These two modules will be explained further shortly after. Input images are processed by convolutional layers and concatenated with the per-step event features extracted from the read network via latent fusion. The fused features are processed and fed to another convolutional LSTM to propagate temporal knowledge along time. Finally, a decoder takes the joint features and generates the deblurred image.

\textbf{Motion compensation.} We use a motion compensation module to warp previous deblurring result $\mathcal{I}_{i+1}$ and generate an initialization of the $i$th time step. Although Eqn.~\eqref{eq:event_update2} achieves this by event integration, we find it more effective to predict a flow field from which we directly warp the clean result $\mathcal{I}_{i+1}$ as additional guidance. Motion compensation for events have already been discussed in~\cite{GallegoCVPR18}. For efficiency, we adopt a FlowNetS architecture~\cite{DosovitskiyICCV15} to take events $\mathbb{E}_{i \sim {i+1}}$ as input and directly regress forward flows from $i$ to $i+1$. Warping is implemented with a differentiable spatial transformer layer~\cite{LaiECCV18,JiangCVPR18}.

\textbf{Directional event filtering.}
The initial estimates $\hat{\mathcal{I}}_i$ may suffer unaddressed blur due to the naive blurring model~\eqref{eq:solve_init} and the noisiness of events. We alleviate this issue with the aid of sharp boundary prior, a widely explored image prior for blind deblurring~\cite{ChoTOG09,XuECCV10}, extracted from events $\mathbb{E}_{i \sim i+1}$.

\begin{figure}[!t]
	\centering
	\includegraphics[width=1.0\columnwidth]{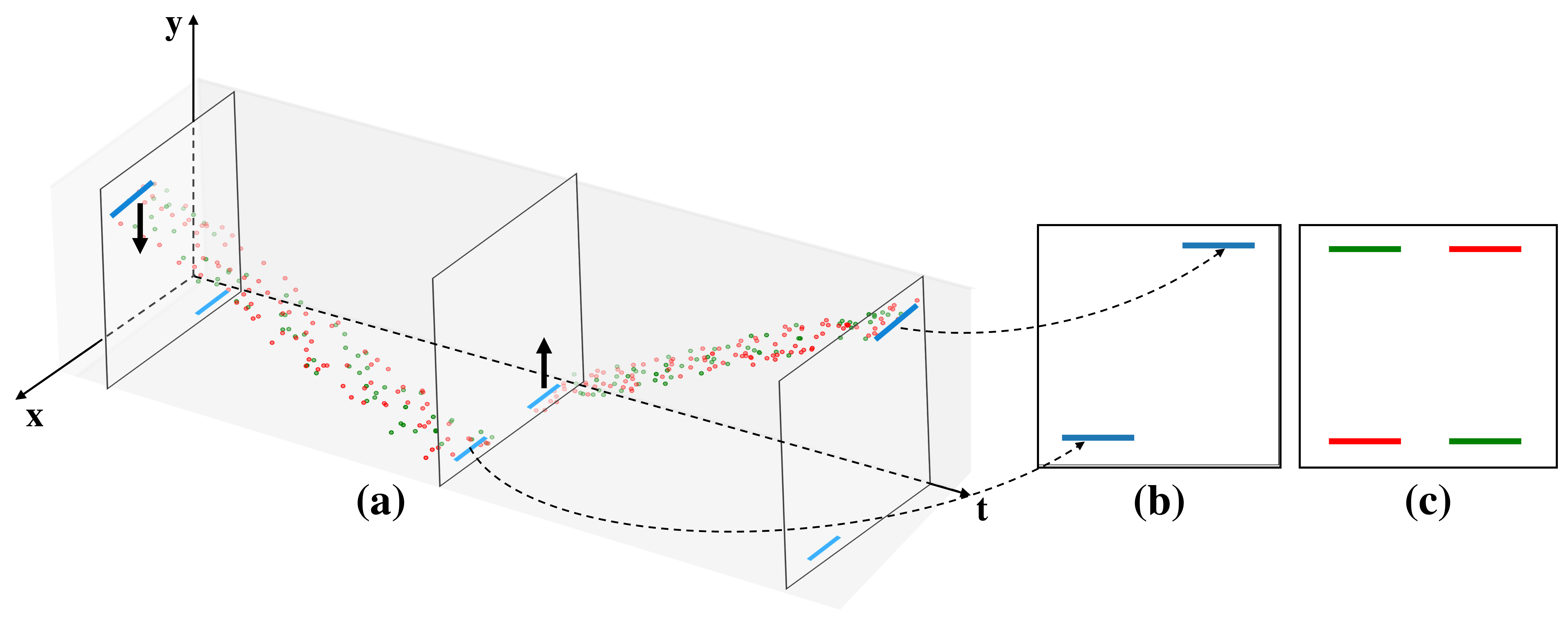}
	\caption{Motivation of adaptive event sampling. (a) A toy scene where the top line moves down first, after which the bottom line moves up. Events with positive and negative polarities are shown as red and green dots, respectively. (b) The projected image of the scene after the imaging process. Scene boundaries correspond to the latest triggered events, which may vary for different positions, as indicated by arrows. (c) The accumulation map of events.} 
	\label{fig:taca_mot}
\end{figure}

Events indicate local change of scene illuminace and reveal physical boundaries. However, as scene boundaries are moving, at a specific time they are only spatially aligned with the \textit{latest} events triggered at their positions. As a toy example, Fig.~\eqref{fig:taca_mot} shows after the imaging the top and bottom lines correspond to events at two different time points. It gives that one can generate scene boundary prior by sampling events at proper space-time positions. Note that due to variation of scene depth, different scene parts may have distinct motion, and position-adaptive sampling is essential.

Besides, as events are sparse, noisy, and non-uniformly distributed signals, a robust sampling process should decide both where (\ie~center) and how many (\ie~scale) to sample. We learn this task from data via differentiable sampling and filtering. For each image position $\bm{\mathrm{p}}$, a temporal center $c \left( \bm{\mathrm{p}} \right)$ and a set of $2k+1$ filtering coefficients $ \{ \alpha_i \}_{i=-k}^k$, where $k$ is the support of filtering kernel, are predicted with a small network from the events, satisfying $\forall i, \alpha_i \geq 0$ and  $\sum_{i=-k}^k \alpha_k = 1$. The filtered result is obtained by
\begin{equation}
\mathcal{G} \left( \bm{\mathrm{p}} \right) = \sum_{i=-k}^k \alpha_k s \left( \bm{\mathrm{p}} + \lambda k \bm{\mathrm{d}} \left( \bm{\mathrm{p}}, c \left( \bm{\mathrm{p}} \right) \right), c \left( \bm{\mathrm{p}} \right) + \lambda k \right),
\end{equation}
where $\lambda$ defines sampling stride (we use $k=2$, $\lambda=1$), $s \left( \cdot, \cdot \right)$ denotes a sampling function in space-time domain. For the stacked event frame representation of events $\mathbb{E}_{i}^{i+1}$, one can apply the trilinear kernel for continuous sampling~\cite{LiuZICCV17}. Note that the velocity $\bm{\mathrm{d}}$ should follow the direction of local motion of events at space-time point $\left( \bm{\mathrm{p}}, c \left( \bm{\mathrm{p}} \right) \right)$ to filter along the density surface of events but not across it.

To get local velocity, we reuse the flow vectors predicted by motion compensation module. We assume object velocity stays constant, which is roughly true in this context as there is just a fraction of time duration (\ie~only $1/\left( T-1 \right)$ the exposure). Motion compensation gives the velocities of all the positions $\bm{\mathrm{p}}_0 \in \mathbb{P}$ at time $i$, $\bm{\mathrm{d}} \left( \bm{\mathrm{p}}_0, i \right)$. At time $c \left( \bm{\mathrm{p}} \right)$, a pixel $ \bm{\mathrm{p}}_0$ would be shifted by the flows to a new position:

\begin{equation}
\bm{\mathrm{n}} \left( \bm{\mathrm{p}}_0 \right) = \bm{\mathrm{p}}_0 + \left( c \left( \bm{\mathrm{p}} \right) - i \right) \bm{\mathrm{d}} \left( \bm{\mathrm{p}}_0, i \right).
\end{equation}
Note that $\bm{\mathrm{n}} \left( \bm{\mathrm{p}}_0 \right)$ inherits the velocity of $\bm{\mathrm{p}}_0$ under the local constancy assumption: $\bm{\mathrm{d}} \left( \bm{\mathrm{n}} \left( \bm{\mathrm{p}}_0 \right), c \left( \bm{\mathrm{p}} \right) \right) = \bm{\mathrm{d}} \left( \bm{\mathrm{p}}_0, i \right)$.

However, the intersected positions at time plane $c \left( \bm{\mathrm{p}} \right)$, namely $\{ \bm{\mathrm{n}} \left( \bm{\mathrm{p}}_0 \right) | \bm{\mathrm{p}}_0 \in \mathbb{P} \}$, does not ensure complete sampling of the image space. Thus, we resample the velocity at a given target $\bm{\mathrm{p}}$ with a Nadaraya-Watson estimator~\cite{NWEstimiator}:
\begin{equation}
\bm{\mathrm{d}} \left( \bm{\mathrm{p}}, c \left( \bm{\mathrm{p}} \right) \right) = \frac{\sum_{\bm{\mathrm{p}}_0 \in \mathbb{P}} \kappa \left( \bm{\mathrm{n}} \left( \bm{\mathrm{p}}_0 \right) - \bm{\mathrm{p}} \right) \bm{\mathrm{d}} \left( \bm{\mathrm{n}} \left( \bm{\mathrm{p}}_0 \right), c \left( \bm{\mathrm{p}} \right) \right) }{\sum_{\bm{\mathrm{p}}_0 \in \mathbb{P}} \kappa \left( \bm{\mathrm{n}} \left( \bm{\mathrm{p}}_0 \right) - \bm{\mathrm{p}} \right)},
\label{eq:NWE}
\end{equation}
where the kernel $\kappa$ is simply defined with a standard Gaussian. This in spirit shares similarity with the ``gather'' approach in computer graphics for surface rendering~\cite{YuTOG2013}.

Eqn.~\eqref{eq:NWE} uses all $\bm{\mathrm{p}}_0$s to estimate each position $\bm{\mathrm{p}}$, which is inefficient. In practice we only use samples located within a local $L \times L$ window centered around $\bm{\mathrm{p}}$. The window size $L$ should account for the maximal spatial displacement of pixels, which we find $L = 20$ sufficient. All of the proposed steps are differentiable, and can be plugged into the network for end-to-end training.

\textbf{Loss Function}.
We use the following joint loss function: 
\begin{equation}
\mathcal{L}_{total} = \mathcal{L}_{content} + \lambda_{a} \mathcal{L}_{adv} + \mathcal{L}_{flow} + \lambda_t \mathcal{L}_{tv},
\end{equation}
Here, $\mathcal{L}_{content}$ is the photometric $\ell_1$ loss $\frac{1}{T}\sum_{i=1}^T \lVert \mathcal{I}^*_i - \mathcal{I}^g_i \rVert $, where $\mathcal{I}^g_i$ is the groundtruth clean image. To improve sharpness of the result, we also incorporate an adversarial loss $\mathcal{L}_{adv}$. We use the same PatchGAN discriminator~\cite{IsolaCVPR17} and follow its original loss definitions strictly.

The flow network introduces two other loss terms. The first $\mathcal{L}_{flow}$ is the photometric reconstruction loss: 
\begin{equation}
\mathcal{L}_{flow} = \frac{1}{T-1} \sum_{i=1}^{T-1} \lVert \omega \left( \mathcal{I}^*_{i+1}, \mathcal{F}_{i \rightarrow i+1} \right) - \mathcal{I}^g_i \rVert,
\end{equation}
where $\omega \left( \cdot, \cdot \right)$ is a backward warping function using forward flows $\mathcal{F}_{i \rightarrow i+1}$, and $\mathcal{L}_{tv} = \frac{1}{T-1} \sum_{i=1}^{T-1} \lVert \nabla \mathcal{F}_{i \rightarrow i+1} \rVert$ is the total variation loss for flow field smoothing. For these terms, we follow the same definitions of~\cite{JiangCVPR18}. The weights $\lambda_a$ and $\lambda_t$ are set to $0.01$ and $0.05$, respectively.

\begin{table*}[t!]
	\centering
	\small
	\caption{Single-image motion deblurring performance on the GoPro dataset.}
	\begin{tabular}{cccccccccccc}
		\toprule
		Models & DCP~\cite{PanCVPR16} & MBR~\cite{SunCVPR15} & FLO~\cite{GongCVPR17} & EVS~\cite{JinCVPR18} & SRN~\cite{TaoCVPR18} & SVR~\cite{ZhangCVPR18} & DMS~\cite{NahCVPR17} & MPN~\cite{ZhangCVPR19} & BHA~\cite{PanCVPR19} & Ours \\
		\midrule
		PSNR & 23.50 & 25.30 & 26.05 & 26.98 & 30.26 & 29.18 & 29.08 & 31.50 & 29.06 & \textbf{31.79} \\
		SSIM & 0.834 & 0.851 & 0.863 & 0.892 & 0.934 & 0.931 & 0.914 & 0.948 & 0.943 & \textbf{0.949} \\
		\bottomrule
		\label{tab:gopro_image_metrics}
	\end{tabular}
	\vspace{-5mm}
\end{table*}

\begin{figure*}
	\centering
	\includegraphics[width=1.0\textwidth]{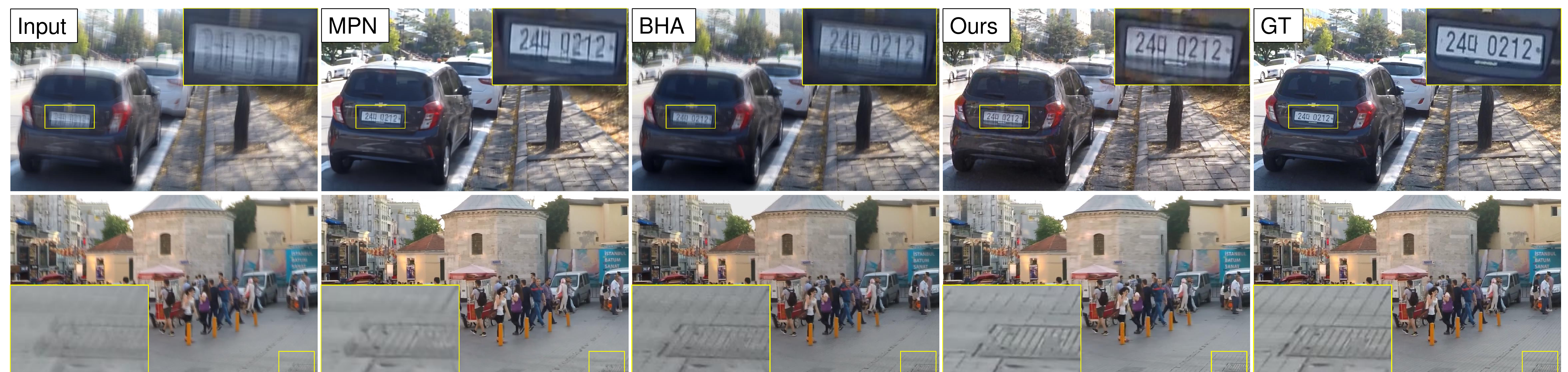}
	\caption{Visual comparisons on the GoPro dataset. From left to right, we show two examples with the blurred image, results of MPN~\cite{ZhangCVPR19}, BHA~\cite{PanCVPR19} and our approach, as well as groundtruth sharp image, respectively. Zoom in for better view.} 
	\label{fig:gopro_results}
	\vspace{-3mm}
\end{figure*}

\section{Experiments}
\subsection{Experimental Settings} 
\textbf{Dataset preparation.} We use two datasets for evaluation. First, we evaluate on the GoPro~\cite{NahCVPR17} dataset which is widely adopted for image motion deblurring and recently used by~\cite{PanCVPR19} to benchmark event-based deblurring. To synthesize events reliably, we use the open ESIM event simulator~\cite{RebecqCORL18}. We follow the suggested training and testing split. The blurred image is also provided officially by averaging nearby (the number varies from $7$ to $13$) frames.

As there lacks a large-scale dataset for evaluating event-based motion deblurring in real-world scenarios, we capture a novel dataset of urban environment, called \textit{Blur-DVS}, with a DAVIS240C camera. It hybrids a high speed event sensor with a low frame-rate Active Pixel Sensor (APS) recording intensity images at $180 \times 240$ . Thus, APS may suffer motion blur in fast moving. We collect two subsets for evaluation. The \textit{slow} subset consists of $15246$ images captured with slow and stable camera movement of relatively static scenes, thus motion blur rarely happens. We synthesize motion blurs by averaging nearby $7$ frames, resulting into $2178$ pairs with blurred image and sharp sequence. In this manner, we can conduct quantitative benchmarkings. We select $1782$ pairs for training, and $396$ for testing. The \textit{fast} subset consists of additional $8$ sequences with $740$ frames in total, captured under fast camera movement of fast moving scenes to investigate how the proposed approach generalizes to real motion blur. However, there is no groundtruth data available on this subset.

\begin{table}[t!]
	\centering
	\small
	\caption{Video reconstruction performance on the GoPro dataset.}
	\begin{tabularx}{\columnwidth}{p{0.95cm}<{\centering}p{0.95cm}<{\centering}p{1.2cm}<{\centering}p{0.95cm}<{\centering}p{0.95cm}<{\centering}p{1cm}<{\centering}}
		\toprule
		Models & CIE~\cite{ScheerlinckACCV18} & CIE+SRN$^*$ & EVS~\cite{JinCVPR18} & BHA~\cite{PanCVPR19} & Ours \\
		\midrule
		PSNR & 25.84 & 26.34 & 25.62 & 28.49 & \textbf{29.67} \\
		SSIM & 0.790 & 0.809 & 0.856 & 0.920 & \textbf{0.927} \\
		\bottomrule
		\label{tab:gopro_video_metrics}
	\end{tabularx}
	\vspace{-7mm}
	\begin{tablenotes}
		\item *A hybird baseline that adopts CIE to reconstruct images first, then SRN to deblur each image. See~\cite{PanCVPR19} for details.
	\end{tablenotes}
	\vspace{-6mm}
\end{table}

\textbf{Method comparison.} We conduct extensive comparisons with recent motion deblurring methods with available results and/or codes. They include image-based methods: DCP~\cite{PanCVPR16}, MBR~\cite{SunCVPR15}, FLO~\cite{GongCVPR17}, DMS~\cite{NahCVPR17}, EVS~\cite{JinCVPR18}, SRN~\cite{TaoCVPR18}, SVR~\cite{ZhangCVPR18} and MPN~\cite{ZhangCVPR19}, and the state-of-the-art event-based motion deblurring method BHA~\cite{PanCVPR19}. We also compare with three event-based video reconstruction methods, including CIE~\cite{ScheerlinckACCV18}, MRL~\cite{MundaIJCV18} and the state-of-the-art learning-based approach ETV~\cite{RebecqCVPR19}. PSNR and SSIM metrics are used for quantitative evaluation. 

\textbf{Implementation details.} For both datasets, our training adopts a batch size of $2$ training pairs and Adam optimizer. The network is trained for $400$ epochs, with a learning rate $10^{-4}$ at the beginning and linearly decayed to zero starting from the $200$th epoch. All the components of the network are jointly trained from scratch.

\subsection{Comparisons with State-of-the-Art Models}
On the GoPro dataset, we report the results on both single image deblurring (\ie~only recovering the middle frame) and video reconstruction (\ie~recover all the sharp frames) in Table~\ref{tab:gopro_image_metrics} and~\ref{tab:gopro_video_metrics}, respectively. Numbers of other approaches are directly taken from papers. Our approach achieves the top place in both tasks, demonstrating the advantages of event-assisted deblurring than purely relying on images, and the superiority of the proposed framework over physical reconstruction model. We show visual comparisons on two fast moving scenes in Fig.~\ref{fig:gopro_results}: while image-based method MPN cannot well address such blur, BHA is sensitive to the noise of events especially along object edges. Our approach generates cleaner and sharper results.

\begin{table*}[t!]
	\centering
	\small
	\caption{Single-image deblurring performance on the Blur-DVS dataset.}
	\begin{tabularx}{\textwidth}{p{1.05cm}<{\centering}p{1.05cm}<{\centering}p{1.05cm}<{\centering}p{1.05cm}<{\centering}p{1.05cm}<{\centering}p{1.05cm}<{\centering}p{1.05cm}<{\centering}p{1.05cm}<{\centering}p{1.05cm}<{\centering}p{1.05cm}<{\centering}p{1.05cm}<{\centering}p{1.05cm}<{\centering}}
		\toprule
		Models & DMS~\cite{NahCVPR17} & SRN~\cite{TaoCVPR18} & SRN+$^*$ & MPN~\cite{ZhangCVPR19} & MPN+$^*$  & CIE~\cite{ScheerlinckACCV18} & MRL~\cite{MundaIJCV18} & ETV~\cite{RebecqCVPR19} & ETV+$^*$ & BHA~\cite{PanCVPR19} & Ours \\
		\midrule
		PSNR & 20.48 & 20.21 & 24.92 & 23.52 & 26.08 & 19.02 & 10.59 & 16.89 & 24.81 &  22.43 & \textbf{26.48} \\
		SSIM & 0.572 & 0.567 & 0.821 & 0.753 & 0.831 & 0.478 & 0.195  & 0.597 & 0.790 & 0.715 & \textbf{0.839}  \\
		\bottomrule
		\label{tab:dvs_image_metrics}
	\end{tabularx}
	\vspace{-7mm}
	\begin{tablenotes}
		\item *SRN+, MPN+ and ETV+ denote enhanced versions of SRN, MPN, ETV respectively. See text for details.
	\end{tablenotes}
	\vspace{-3mm}
\end{table*}

\begin{figure*}[t!]
	\centering
	\includegraphics[width=1.0\textwidth]{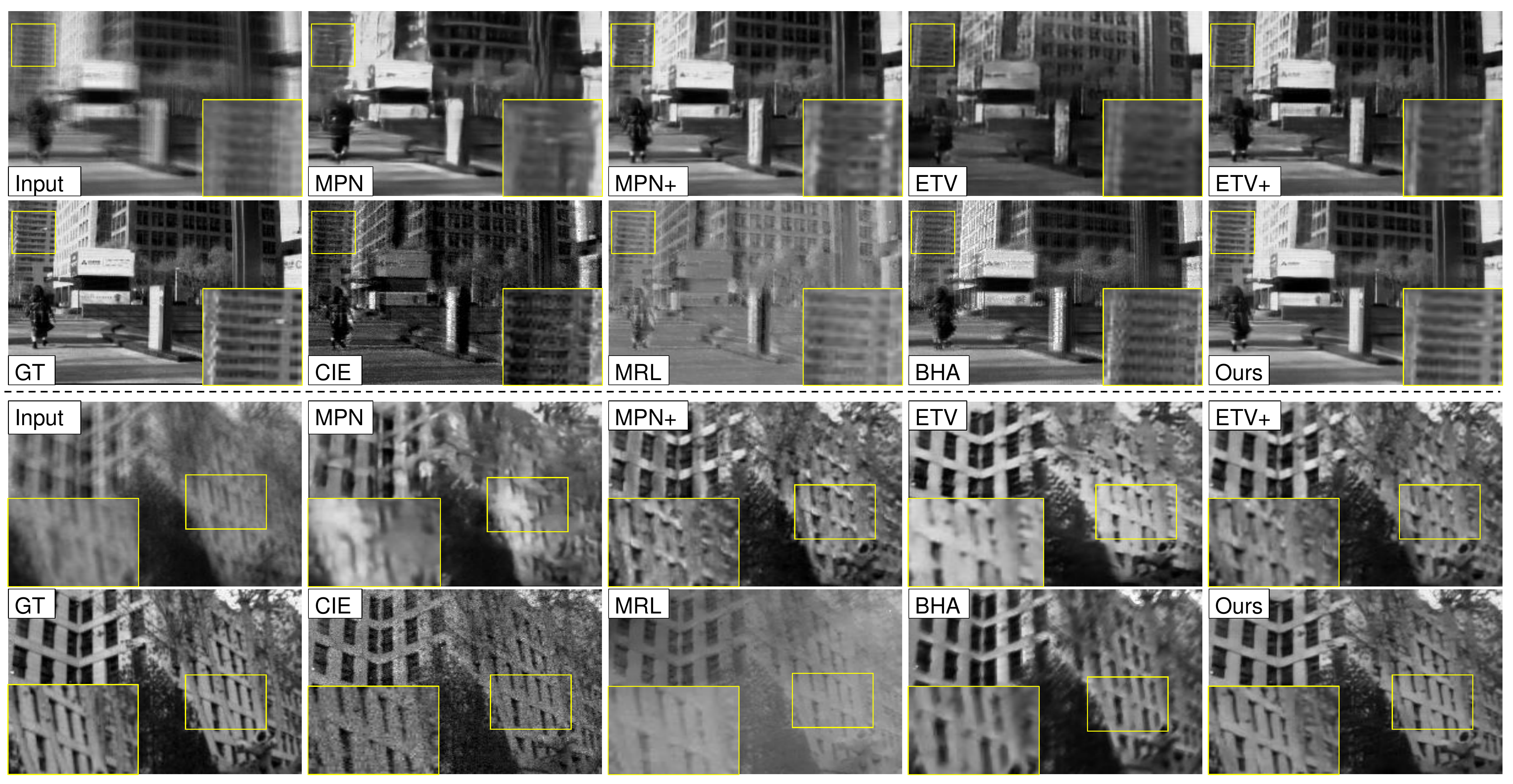}
	\caption{Representative results of two examples generated by different approaches on the \textit{slow} subset of Blur-DVS dataset. More results can be found in our supplementary material. Zoom in for better view.} 
	\label{fig:dvs_image_results}
	\vspace{-2mm}
\end{figure*}

\begin{table}[t!]
	\centering
	\small
	\caption{Video reconstruction performance on Blur-DVS dataset.}
	\begin{tabularx}{\columnwidth}{p{0.65cm}<{\centering}p{0.8cm}<{\centering}p{0.95cm}<{\centering}p{0.95cm}<{\centering}p{0.5cm}<{\centering}p{1.0cm}<{\centering}p{0.7cm}<{\centering}p{0.6cm}<{\centering}}
		\toprule
		Models & CIE~\cite{ScheerlinckACCV18} & MRL~\cite{MundaIJCV18} & ETV~\cite{RebecqCVPR19} & ETV+ & BHA~\cite{PanCVPR19} & Ours \\
		\midrule
		PSNR & 18.94 & 10.57 & 16.60 & 24.10 & 22.06 & \textbf{25.33} \\
		SSIM & 0.473 & 0.194 & 0.587 & 0.777 & 0.699 & \textbf{0.827} \\
		\bottomrule
		\label{tab:dvs_video_metrics}
	\end{tabularx}
	\vspace{-10mm}
\end{table}

Note that GoPro dataset mainly presents small to moderate motion blur, thus the blurred input is of good quality and improvement from events is marginal. Thus recent powerful architectures SRN and MPN get very promising results though they do not see events. For this reason, we compare our approach with state-of-the-art methods on the proposed Blur-DVS dataset, in which severe motion blur are more universal. Again, we report results on single image deblurring (Table~\ref{tab:dvs_image_metrics}) and video reconstruction (Table~\ref{tab:dvs_video_metrics}) tasks. Note that for fair comparisons, The learning-based methods SRN, MPN and ETV are finetuned on the training set of Blur-DVS. We also compare with their enhanced versions that see both image and events: for image-based methods SRN and MPN, we concatenate the input blurred image with all the $48$ ($8$ binned frames in each time interval and $(7-1)$ intervals) event frames. For the event-based method ETV, we also feed the blurred image along with the events to each of its recurrent reconstruction step. We denote these variants as SRN+, MPN+ and ETV+, respectively.

In Table~\ref{tab:dvs_image_metrics} and~\ref{tab:dvs_video_metrics}, the proposed approach achieves the best results. It also outperforms all the enhanced variants, demonstrating the effectiveness of the proposed framework. Fig.~\ref{fig:dvs_image_results} illustrates that: 1) in case of fast motion, image-based cues alone are not sufficient, limiting performance of MPN; 2) the physical model BHA is prone to noise and presents unaddressed blur due to the lossy sampling mechanism of events; 3) event-based reconstruction methods CIE, MRL and ETV do not restore scene contrast correctly due to the lack of image guidance and/or the simplified physical model. Our approach does not suffer the mentioned issues, and presents sharper results even than the enhanced image+event variants equipped with powerful architectures.

\begin{figure*}[t!]
	\centering
	\includegraphics[width=1.0\textwidth]{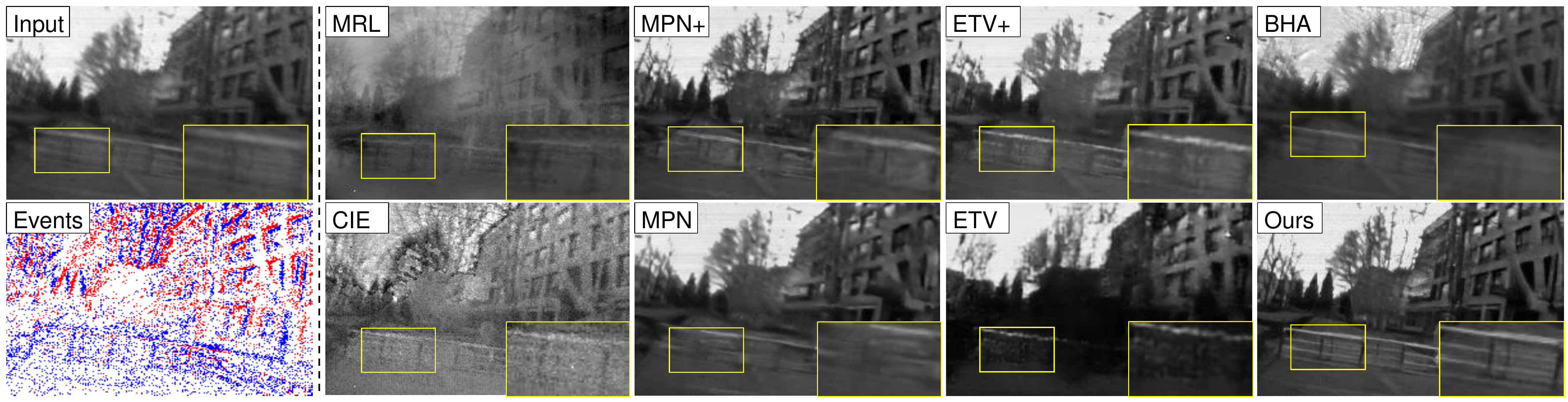}
	\caption{Representative results generated by different approaches on the \textit{fast} subset (real-world motion blur) of Blur-DVS dataset. More results can be found in our supplementary material. Zoom in for better view.} 
	\label{fig:dvs_real_results}
	\vspace{-7mm}
\end{figure*}

Finally, we analyse the generalization behavior to real-world motion blur. As shown in Fig.~\ref{fig:dvs_real_results}, the proposed approach achieves the best visual quality. We suspect that the explicit modeling of motion deblurring and introduction of strong deblurring priors may alleviate the learning difficulty and avoid potential overfitting in more black-box architectures. In practice we find such improvement consistent on real data, as demonstrated by more results on the \textit{fast} subset provided in our supplementary material.

\subsection{Performance Analysis}
\begin{figure}
	\centering
	\includegraphics[width=0.99\columnwidth]{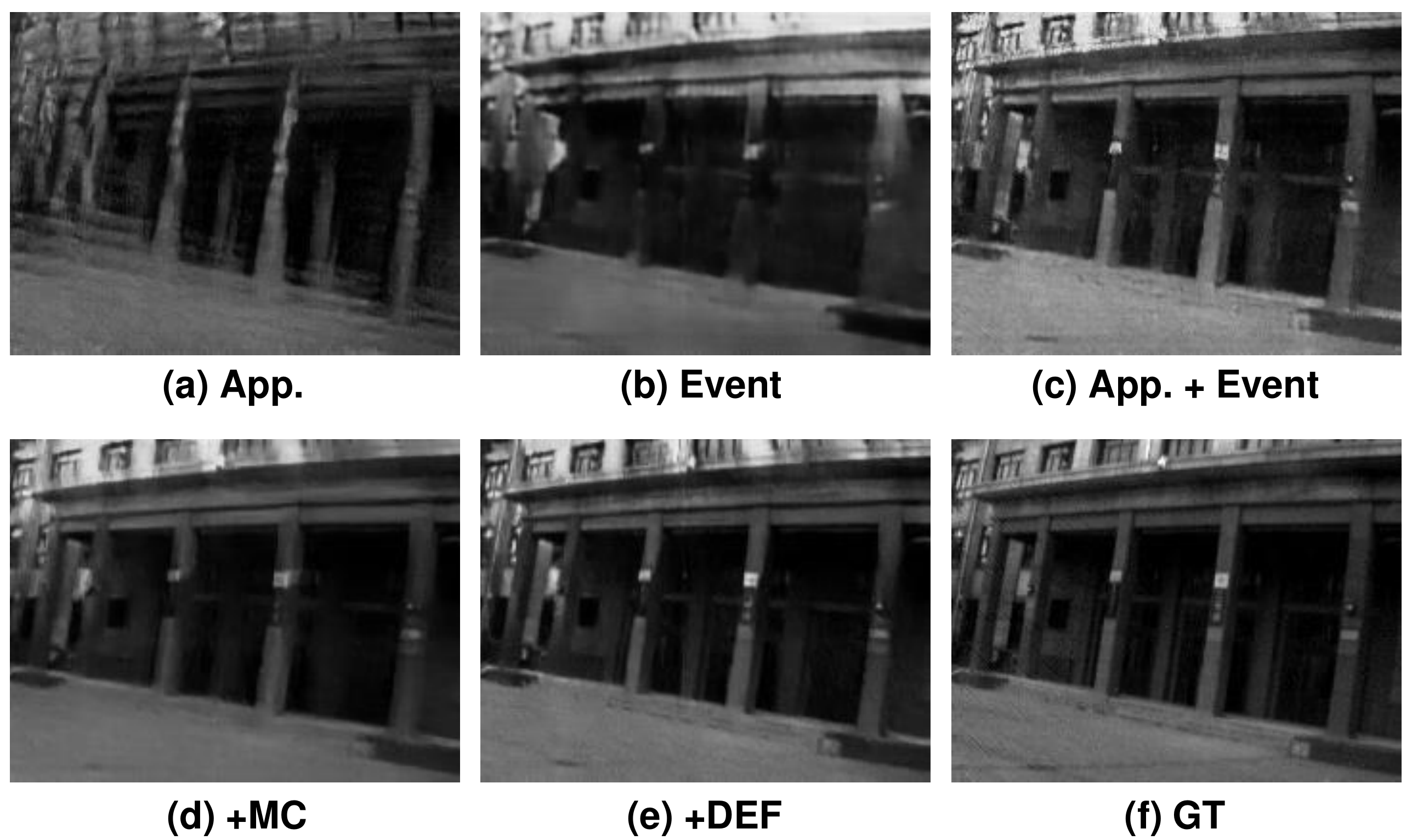}
	\caption{Visually analysing the contributions of different components on the DVS-Blur dataset. See text for details. } 
	\label{fig:ablation_study}
\end{figure}

\textbf{Analysing different components.} We isolate the important algorithm components to see ther contributions to the final performance, and summarize the results in Table~\ref{tab:ablation_study} and Fig.~\ref{fig:ablation_study}. As it shows, each component is necessary to improve the PSNR and SSIM of the results. Using image appearance only without events (App.) cannot deblur the image well. Using events only, on the other hand, recovers plenty of details but intensity contrast is not well recovered (see Fig.~\ref{fig:ablation_study} (b)). Using both input signals (App. + event) achieves better results, but the reconstructed image is not very smooth due to noise (\eg~the ground in Fig.~\ref{fig:ablation_study} (c)). Further incorporating motion compensation (+MC) helps in these aspects as it imposes temporal smoothness. Finally, further introducing the directional event filtering module (+DEF), sharper results and richer details can be generated thanks to the learned boundary guidance.

\begin{table}[t!]
	\centering
	\small
	\caption{Component analysis on the Blur-DVS dataset. ``App.'' and ``event'' denotes using the blurred image appearance and event data as input, respectively. ``MC'' and ``DEF'' refer to the motion compensation and directional event filtering modules, respectively.}
	\begin{tabular}{ccccccc}
		\toprule
		App. & Event & +MC & +DEF & PSNR & SSIM \\
		\midrule
		\cmark & \xmark & \xmark & \xmark & 16.50 & 0.418 \\
		\xmark & \cmark & \cmark  & \xmark  & 16.38 & 0.560 \\
		\cmark & \cmark & \xmark  & \xmark  & 23.39 & 0.760 \\
		\cmark & \cmark & \cmark  & \xmark  & 24.71 & 0.786 \\
		\cmark & \cmark & \cmark  & \cmark  & 25.33 & 0.827 \\
		\bottomrule
		\label{tab:ablation_study}
	\end{tabular}
	\vspace{-8mm}
\end{table}

\textbf{Justification of the DEF module.} In Table~\ref{tab:def}, we justify the necessity of the proposed directional event filtering module. Here, ``w/o guid.'' does not include boundary guidance in the whole pipeline. On the contrary, ``guid only.'' discards event features in each sequential deblurring step while using boundary guidance only as additional cue. We further design a variant ``+param.'', which does not incorporate DEF but has additional convolution layers in the encoder of \textit{process} network which exceeds the current parameter size. Results show that the learned boundary guidance greatly improves the estimation (from $0.786$ to $0.827$ in SSIM), and itself without other cues can already leads to promising results. Simply enlarging the network size, however, does not observe meaningful improvement.

\begin{table}[t!]
	\centering
	\small
	\caption{Analysing the directional event filtering module on the DVS-Blur dataset. See text for details.}
	\begin{tabular}{ccccc}
		\toprule
		Models & guid. only & w/o guid. & full & +param. \\
		\midrule
		PSNR & 25.16 & 24.71 & 25.33 & 24.64 \\
		SSIM & 0.816 & 0.786 & 0.827 & 0.788 \\
		\bottomrule
		\label{tab:def}
	\end{tabular}
	\vspace{-10mm}
\end{table}

\begin{figure}
	\centering
	\includegraphics[width=1.0\columnwidth]{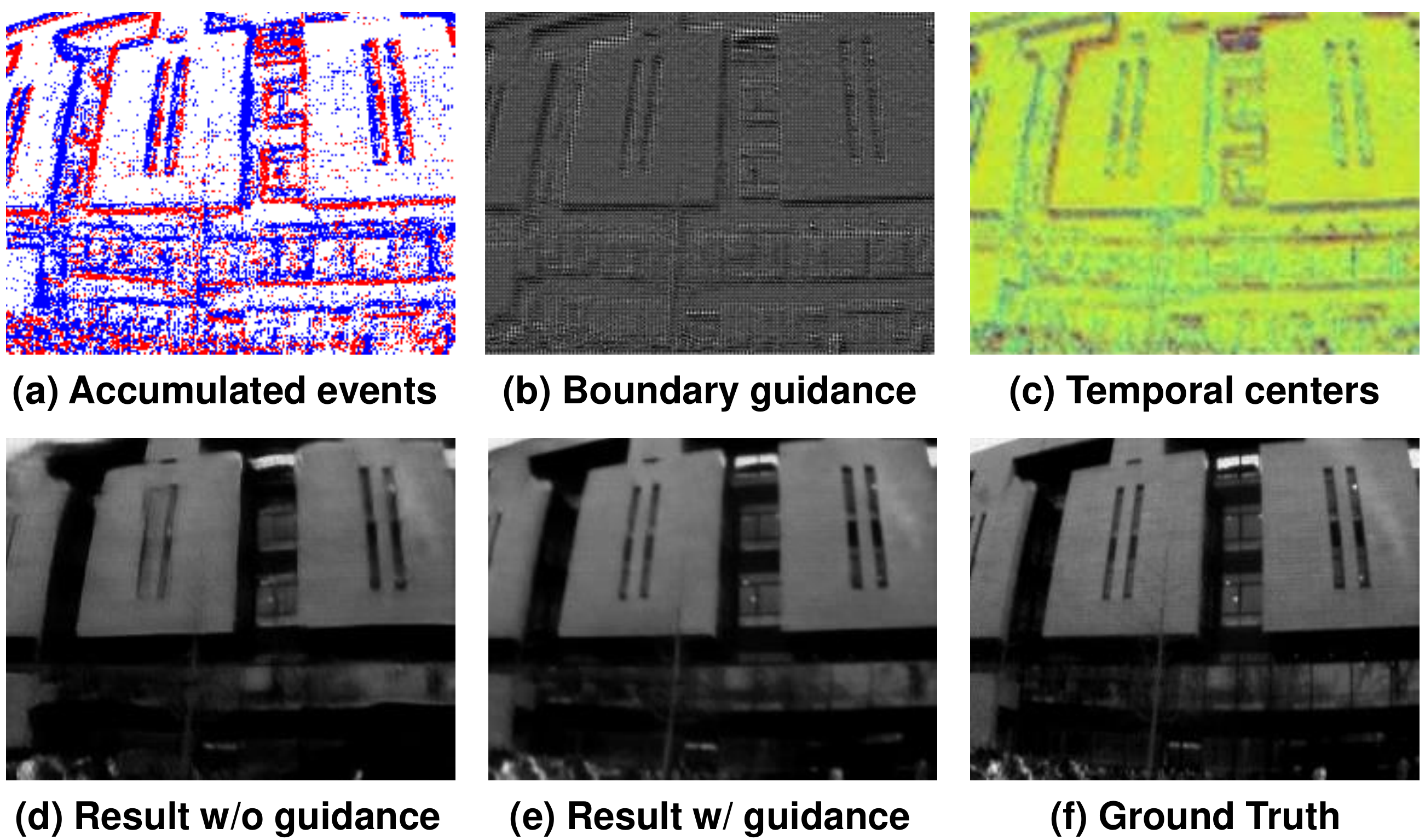}
	\caption{Visualizing learned boundary guidance. Note how motion boundaries from different time stamps are selected in the attention map (c) (red for large value and blue for small values).} 
	\label{fig:attention}
	\vspace{-3mm}
\end{figure}

In Fig.~\ref{fig:attention}, we visualize the impact of learned boundary guidance. Note how the network learns to select different time centers according to the scene's motion (Fig.~\ref{fig:attention} (c)). Boundary guidance improves the sharpness of the scene significantly and recovers missing details (Fig.~\ref{fig:attention} (e) and (f)).

\begin{figure}[t!]
	\centering
	\includegraphics[width=1.0\columnwidth]{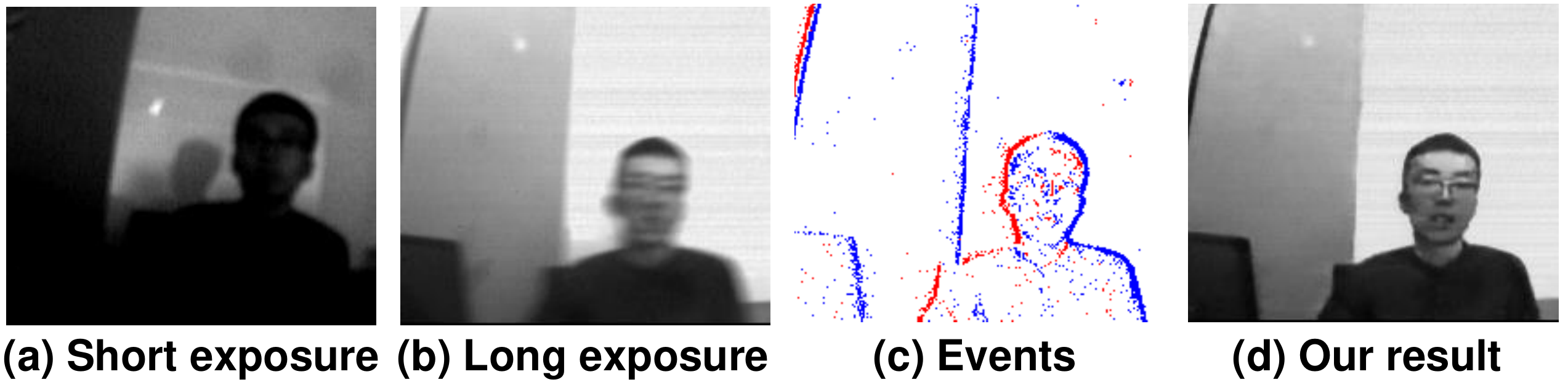}
	\caption{Low-light photography using our approach. Images and events are captured with DAVIS240C camera in an indoor scene.} 
	\label{fig:light-starved}
	\vspace{-5mm}
\end{figure}

\textbf{Low-light photography.} A potential application of the proposed approach is low-light photography, as shown in Fig~\ref{fig:light-starved}. The short-exposure ($13$ms) image is light-starved. The long-exposure ($104$ms) one, however, may suffer severe motion blur. Leveraging event cues, our approach generates natural results without such blur.

\section{Conclusion}
In this work, we propose to extract a video from a severe motion-blurred image under the assistance of events. To this end, a novel deep learning architecture is proposed to effectively fuse appearance and motion cues at both global and local granularity. Furthermore, sharp event boundary guidance is extracted to improve reconstructed details with a novel directional event filtering module. Extensive evaluations show that the proposed approach achieves superior performance than various existing image and event-based methods, on both synthetic and real-world datasets.

\small
\vspace{2mm}
\textbf{Acknowledgements.} We thank the reviewers for their valuable feedback. This work is supported by Beijing Posdoctoral Research Foundation (Grant No. ZZ-2019-89), National Key R\&D Program of China under contract No. 2017YFB1002201, National Natural Science Fund for Distinguished Young Scholar (Grant No. 61625204) and partially supported by the Key Program of National Science Foundation of China (Grant No. 61836006). 

{\small
\bibliographystyle{ieee_fullname}
\bibliography{egbib}
}

\end{document}